%% file: conference_101719.tex
\documentclass[letterpaper, 10 pt, conference]{ieeeconf}
\IEEEoverridecommandlockouts
\usepackage{cite}
\usepackage{amsmath,amssymb,amsfonts}
\usepackage{algorithmic}
\usepackage{graphicx}
\usepackage{textcomp}
\usepackage{xcolor}
\usepackage{siunitx}
\usepackage{float}
\usepackage{subcaption}
\usepackage{booktabs}
\usepackage{siunitx}
\usepackage{balance}
\usepackage{multirow}

\newcolumntype{P}[1]{>{\centering\arraybackslash}p{#1}}
\newcolumntype{M}[1]{>{\centering\arraybackslash}m{#1}}
\def\BibTeX{{\rm B\kern-.05em{\sc i\kern-.025em b}\kern-.08em
    T\kern-.1667em\lower.7ex\hbox{E}\kern-.125emX}}
    
\begin{document}
 
\title{\LARGE \bf EdgeLPR: On the Deep Neural Network trade-off between Precision and Performance in LiDAR Place Recognition \\

\thanks{
*Computational resources provided by computing@unipi, a Computing Service provided by
University of Pisa.}
}
\author{Pierpaolo Serio$^{1}$, Hetian Wang$^{2}$, Zixiang Wei$^{2}$, Vincenzo Infantino$^{3}$, Lorenzo Gentilini$^{3}$,\\Lorenzo Pollini$^{1}$,   Valentina Donzella$^{2}$%
\thanks{$^{1}$Department of Information Engineering, University of Pisa, Pisa, Italy. {\tt\small \{pierpaolo.serio@phd.unipi.it, lorenzo.pollini@unipi.it\}}}%
\thanks{$^{2}$School of Engineering and Material Science, Queen Mary University of London, London, U.K. {\tt\small \{zixiang.wei, hetian.wang, v.donzella\}@qmul.ac.uk}}
\thanks{$^{3}$ Research and Development, Toyota Material Handling Manufacturing Italy, Bologna, Italy. {\tt\small \{vincenzo.infantino, lorenzo.gentilini\}@toyota-industries.eu}}}

\maketitle

\begin{abstract}
Place recognition is essential for long-term autonomous navigation, enabling loop closure and consistent mapping. Although deep learning has improved performance, deploying such models on resource-constrained platforms remains challenging. This work explores efficient LiDAR-based place recognition for EdgeAI by leveraging Bird’s Eye View representations to enable lightweight image-based networks. We benchmark representative architectures without aggregation heads using a unified descriptor scheme based on global pooling and linear projection, and evaluate performance under FP32, FP16, and INT8 quantization. Experiments reveal trade-offs between accuracy, robustness, and efficiency: FP16 matches FP32 with lower cost, while INT8 introduces architecture-dependent degradation. Overall, the presented results are a strong basis for future research on 'use-case'-aware quantisation of Neural Networks for Edge deployment.
\end{abstract}

\section{Introduction}

The rapid progress of Embodied AI is enabling robotic systems to achieve increasingly sophisticated levels of environmental understanding and proprioceptive awareness\cite{billard2025roadmap, casini2026artificial,pisaneschi2026mentalisticinterfaceprobingfolkpsychological}. Within the robotic perception pipeline, Deep Learning has emerged as a key enabler, significantly improving the extraction of meaningful features from sensory data\cite{chen2024role}. These features allow an autonomous agent to construct and continuously refine an internal representation of the environment, which is essential for reliable localization during exploration. A critical capability in this context is the recognition of previously visited locations, commonly referred to as \emph{loop closure}. By detecting revisited places, the system can correct the accumulated drift inherent to long-term state estimation, thereby maintaining the consistency of the constructed map. As such, loop closure constitutes a fundamental component of modern navigation systems. However, increasing performance requirements and the complexity of real-world environments often lead to progressively larger and more complex models. Given the multiplicity of modules in a typical navigation stack (including state estimation, mapping, object detection, segmentation, and place recognition), the concurrent deployment of multiple deep networks can quickly become prohibitive under realistic hardware constraints. This challenge has motivated the development of EdgeAI, which focuses on designing efficient neural architectures that balance accuracy with computational and memory efficiency. In the context of LiDAR-based place recognition, memory consumption represents a particularly critical bottleneck. Traditional approaches require storing and comparing large point clouds over time, resulting in significant storage and computational demands. To address this issue, this work investigates the performance of deep neural networks by using an alternative paradigm based on Bird’s Eye View (BEV) representations derived from point clouds. This transformation enables the use of lightweight image-based models, reducing both the input dimensionality and the memory footprint. Specifically, instead of retaining full point clouds, the system stores compact global descriptors, thereby enabling scalable long-term place retrieval. Within this framework, the main contributions of this paper are as follows.
\begin{itemize}
\item The design and validation of a novel framework for the controlled and reproducible evaluation of lightweight neural networks under different quantization levels applied to place recognition.
\item A comprehensive quantitative assessment of quantisation of state-of-the-art methods, with a focus on performance in complex environments and different precisions.
\item An empirical toolbox to support hardware and performance-informed design choices for next-generation robotic perception systems.
\end{itemize}
Overall, this work aims to provide a principled foundation for the development of resource-efficient, AI-driven place recognition pipelines.



\begin{figure*}[ht]
    \centering
    \includegraphics[width=0.8\linewidth]{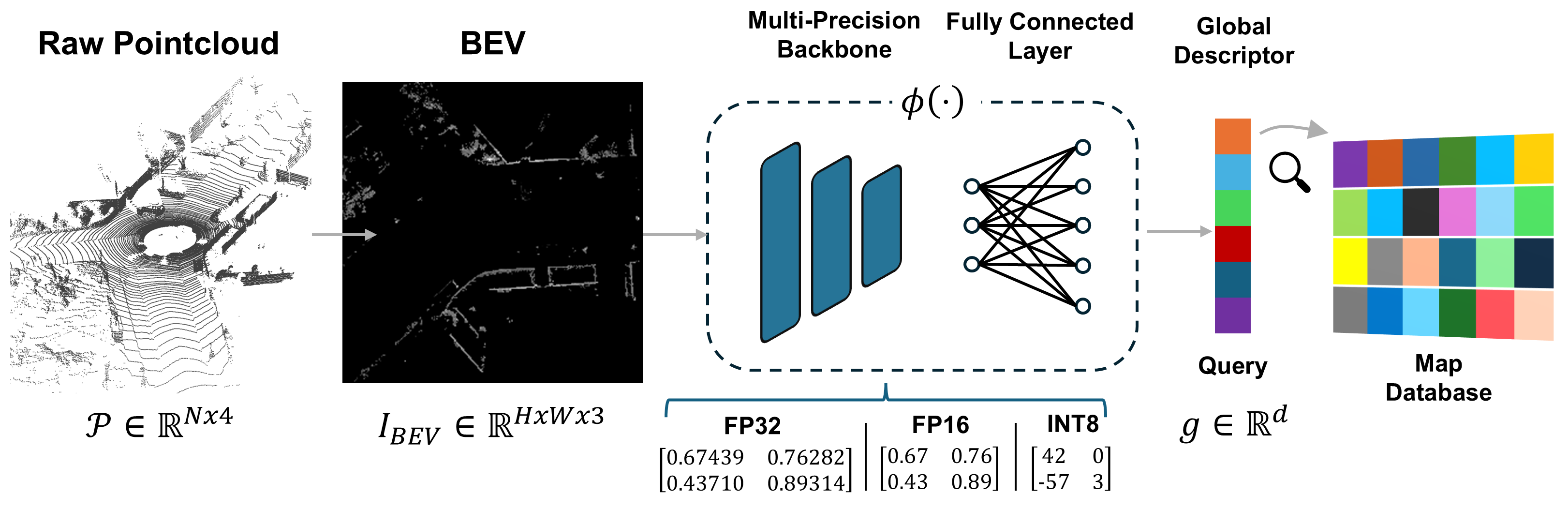}
     \caption{Scheme of the proposed architecture. First, the point cloud is projected in a 256x256 Bird's Eye View format (BEV). Then the Multi-Precision Backbone extracts a meaningful descriptor that is then resized to a desired dimension through a shallow Fully Connected Layer.}
    \label{fig:architecture}
\end{figure*}

\section{Related Work}

Place recognition and loop closure detection are fundamental for long-term autonomy and consistent mapping in robotics and computer vision. Early approaches relied on hand-crafted visual features such as SIFT and SURF~\cite{valgren2010sift}, or BRIEF within bag-of-words frameworks~\cite{galvez2012bags}, enabling scalable retrieval but showing limited robustness to viewpoint, illumination, and environmental changes. The advent of Deep Learning has significantly improved place recognition by enabling the learning of discriminative global descriptors. CNN-based methods, including NetVLAD~\cite{arandjelovic2016netvlad}, aggregate local features into compact representations suitable for large-scale retrieval. Metric learning strategies, such as triplet and contrastive losses, further enhance robustness under challenging conditions. In parallel, LiDAR-based place recognition has gained attention due to its invariance to lighting and robustness in geometrically complex environments~\cite{zhang2024lidar}. Traditional methods rely on geometric descriptors, such as histograms of surface normals~\cite{luo2022lidar} and scan context representations~\cite{kim2018scan, kim2021scan, wang2020intensity}. While efficient, their discriminative power can degrade in environments with repetitive structures. Learning-based approaches have also been extended to LiDAR data. Point-based networks, such as PointNet~\cite{qi2017pointnet}, and transformer-based models~\cite{wu2025sonata}, operate directly on raw point clouds. Although effective, these methods typically require significant computational resources due to the high dimensionality and unordered nature of point cloud data. To mitigate these limitations, several works have proposed transforming point clouds into structured representations\cite{serio2025polar}, such as range images\cite{jung2025imlpr} or Bird’s Eye View (BEV) projections\cite{luo2023bevplace, luo2025bevplace++}, enabling the use of efficient 2D convolutional architectures. The use of BEV representations has proven particularly attractive in the context of resource-constrained systems. By projecting 3D data into a 2D plane, these methods reduce input complexity while preserving spatial relationships relevant for place recognition. This approach facilitates the adoption of lightweight image-based models, originally developed for mobile and embedded vision applications. Nevertheless, the trade-off between efficiency and accuracy remains an open challenge, especially when operating in large-scale or dynamic environments. From a system perspective, the growing complexity of modern robotic pipelines has motivated research into more efficient model design and deployment. Techniques such as model compression, pruning, quantization, and the development of lightweight architectures (e.g., MobileNet\cite{howard2017mobilenets}, ShuffleNet\cite{zhang2018shufflenet}) have been widely investigated under the umbrella of \emph{EdgeAI}. These efforts aim to enable the deployment of deep learning models on platforms with limited computational and memory resources, without significantly compromising performance. Despite these advances, a systematic and controlled analysis of lightweight architectures for LiDAR-based place recognition remains limited in the literature. In particular, the impact of design choices related to input representation, descriptor dimensionality, and model complexity on both performance and resource consumption is not yet fully understood. This gap motivates the present work, which focuses on benchmarking modern lightweight networks and their quantized versions within a unified experimental framework, providing insights into their suitability for real-world robotic applications.

\section{Methodology}

\subsection{Problem Definition}

In this work, place recognition is formulated as a retrieval problem over LiDAR observations. As the robotic platform traverses an environment, each sensor acquisition at time \(t\) is denoted as \(z_t\), corresponding to a LiDAR scan projected into a Bird’s Eye View (BEV) image. Each observation is associated with a spatial pose \(p_t\), and is treated as an instance of a \emph{place}. The map \(\mathcal{M}\) is thus defined as the ordered set of all previously observed frames.

For evaluation purposes, the dataset is partitioned into two disjoint subsets: a \emph{database} \(\mathcal{D}\), containing reference frames, and a set of \emph{queries} \(\mathcal{Q}\), for which correspondences are sought. To avoid trivial matches, query frames are selected with a temporal offset relative to the database.

We model place recognition through a learned encoding function \(f_\theta\), which maps each input observation \(z\) to a compact global descriptor:
\[
g = f_\theta(z) \in \mathbb{R}^d.
\]
Ground-truth correspondences are defined based on spatial proximity. A database element \(j \in \mathcal{D}\) is considered a positive match for a query \(q \in \mathcal{Q}\) if:
\[
\|p_q - p_j\|_2 < \tau,
\]
where \(\tau\) is a predefined distance threshold. Additionally, a temporal constraint \(|t_q - t_j| > \Delta t\) is enforced to prevent trivial matches between consecutive frames. Matches satisfying both spatial and temporal conditions are considered valid place recognitions. A comprehensive view of the proposed pipeline is shown in Figure~\ref{fig:architecture}.

\subsection{Representation and Retrieval Pipeline}
\label{subsection:representation}

The proposed framework adopts a descriptor-based retrieval paradigm tailored for resource-constrained deployment. Each BEV image \(\mathcal{I} \in \mathbb{R}^{H \times W}\) is first adapted to a three-channel format and processed by a lightweight convolutional backbone \(f\), yielding a compact feature representation. In contrast to prior works\cite{luo2025bevplace++,jung2025imlpr} relying on explicit aggregation modules, we directly obtain global descriptors through a combination of spatial pooling and a fully connected projection layer.

Formally, the encoding function is defined as:
\[
g = f_\theta(\mathcal{I}) = \phi\big(\mathrm{pool}(f(\mathcal{I}))\big),
\]
where \(f(\cdot)\) denotes the backbone network, \(\mathrm{pool}(\cdot)\) is an adaptive global pooling operation, and \(\phi(\cdot)\) is a linear projection mapping features to a descriptor space of fixed dimension \(d\). \emph{This design ensures a consistent descriptor size across different architectures, enabling fair comparison.}
Rather than introducing additional aggregation heads, all models share the same descriptor generation strategy, isolating the contribution of the backbone architecture itself. This pipeline significantly reduces both computational and memory requirements. Instead of storing full point clouds, the system maintains only compact global descriptors, enabling scalable long-term operation. Moreover, by leveraging efficient 2D convolutional backbones on BEV projections, the framework avoids the overhead associated with direct 3D point cloud processing, making it particularly suitable for EdgeAI applications. Performance is evaluated using standard retrieval metrics, including Recall@$k$ with $k\in[1,5,10,20]$, which measures the proportion of queries for which the top-ranked database element corresponds to a ground-truth positive. Additional analyses consider the distribution of descriptor distances to characterize robustness under varying matching thresholds.

\subsection{Quantization Strategy}\label{subsection:quantization}

To assess the trade-off between computational efficiency and descriptor quality, we evaluate all backbone architectures under multiple numerical precision regimes, namely FP32, FP16, and INT8 quantization. Given a real-valued tensor \(x \in \mathbb{R}\), its quantized counterpart \(\hat{x}\) is obtained as:
\[
\hat{x} = s \cdot \left( \mathrm{round}\left(\frac{x}{s}\right) + z \right),
\]
where \(s \in \mathbb{R}^+\) is a scaling factor, and \(z \in \mathbb{Z}\) is a zero-point offset. This transformation enables representing activations and weights using low-precision integer formats, thereby reducing memory footprint and improving computational efficiency. 

Focusing on the INT8, we adopt a static post-training quantization (PTQ) strategy~\cite{banner2019post}. In contrast to dynamic quantization, the quantization parameters (scale \(s\) and zero-point \(z\)) are fixed after a calibration phase, thereby avoiding additional overhead during inference. Formally, the quantization process can be interpreted as finding the parameters \(\{s, z\}\) that best approximate the original floating-point activations over a representative calibration set \(\mathcal{X}_{cal}\). This is achieved by minimizing the reconstruction error:
\begin{equation}
    \min_{s, z} \mathbb{E}_{\mathbf{x} \sim \mathcal{X}_{cal}} \left\| \mathbf{x} - \mathcal{Q}(\mathbf{x}; s, z) \right\|_p,
\end{equation}
where \(p=2\) corresponds to the Mean Squared Error (MSE) criterion, and \(\mathcal{Q}(\cdot)\) denotes the quantization operator.  In practice, this calibration phase consists of passing a subset of representative inputs through the network to estimate the dynamic range of activations at each layer. These statistics are then used to determine appropriate scaling factors, ensuring that the discrete quantization levels effectively cover the range of the original signals. As a result, both weights and activations can be converted to INT8 with limited loss of information. From a systems perspective, FP16 reduces memory usage and accelerates computation on compatible hardware (e.g., GPUs) by leveraging half-precision floating-point arithmetic, while INT8 quantization further compresses the model and enables efficient integer-only inference on CPU backends. However, these benefits might come at the cost of reduced numerical precision, which could impact the discriminative power of the learned descriptors. Within the scope of this work, this quantization framework provides a principled way to benchmark lightweight architectures under realistic deployment constraints.


\begin{table*}[ht]
\centering
\caption{Architectural Specifications and Memory Footprint of Evaluated Models}
\label{tab:model_specs}
\begin{tabular}{lcccc}
\toprule
\textbf{Backbone} & \textbf{Total Params} & \textbf{Backbone/Head} & \textbf{Memory (FP32)} & \textbf{Memory (INT8)} \\ 
\midrule
ShuffleNetv2 & 2.28 M & 1.25 M / 1.02 M & 8.69 MB & 2.17 MB \\
MobileNetV3  & 3.93 M & 2.97 M / 0.96 M & 15.00 MB & 3.75 MB \\
ResNet18     & 11.69 M & 11.18 M / 0.51 M & 44.59 MB & 11.15 MB \\
\bottomrule
\end{tabular}
\end{table*}

\section{Experiments}

\subsection{Training}
In contrast to large-scale transformer-based models, our approach focuses on well-established efficient CNN backbones, namely MobileNetV3, ShuffleNetV2, and ResNet18, augmented with a simple fully connected projection layer to produce fixed-dimensional global descriptors and to enable fair comparison. The model's choice aims to represent complementary and well-known design paradigms in lightweight network design. All backbone networks are initialized with ImageNet-pretrained weights and fine-tuned end-to-end for the place recognition task. Given the relatively small capacity of the selected architectures, we do not freeze any layers, allowing the full network to adapt to the BEV domain. The final descriptor is obtained by applying global average pooling followed by a linear projection layer, which maps the feature vector to a shared embedding space of dimension \(d\). The optimization is performed using the AdamW optimizer, which decouples weight decay from gradient updates. A single learning rate is adopted for the entire network, set to \(1\times10^{-3}\), with weight decay \(1\times10^{-4}\). This choice reflects the absence of separate aggregation modules and ensures stable convergence across all evaluated architectures. To learn meaningful representations, we adopt a triplet loss objective. Each training sample consists of an anchor frame, a positive frame corresponding to the same place, and a set of negative frames representing different locations. The loss enforces that descriptors of spatially close observations are pulled together in the embedding space, while those of distant locations are pushed apart. Positive and negative samples are selected based on ground-truth pose information: positives satisfy \(\|p_a - p_p\|_2 < \tau_p\), while negatives satisfy \(\|p_a - p_n\|_2 > \tau_n\), with $\tau_p = 5m$  $\tau_n=10m$ as reported in Table \ref{tab:training_settings}. Training is conducted on $'00'$ sequence of the standard KITTI dataset, where BEV images are generated from raw LiDAR point clouds using intensity as pixel value. To perform the training, the first 3000 BEVs are used as \emph{database} and the remaining as \emph{query}. To improve robustness to viewpoint variations, we apply data augmentation in the form of random in-plane rotations sampled uniformly from \([0, 2\pi)\). Each mini-batch includes multiple negatives per anchor, enabling effective metric learning without requiring excessively large batch sizes. The models are trained for a fixed number of epochs ($50$), and performance is periodically evaluated on a held-out validation split to monitor convergence. To ensure reproducibility, all experiments are conducted with fixed random seeds across PyTorch and NumPy. 

\begin{table}[t]
\centering
\caption{Summary of training configuration.}
\label{tab:training_settings}
\begin{tabular}{ll}
\hline
Optimizer & AdamW \\
Learning Rate & $1\times10^{-3}$\\
Weight Decay & $1\times10^{-4}$ \\
Loss Function & Triplet loss \\
Epochs & $50$ \\
Augmentation & Random rotation $[0,2\pi)$ \\
Positive / Negative dist. & $\tau_p = 5m$ / $\tau_n = 10m$ \\
\hline
\end{tabular}
\end{table}

\subsection{Testing}

We evaluate the proposed models using a unified retrieval pipeline where $\ell_2$-normalized global descriptors $\mathcal{G}$ are indexed via FAISS for efficient nearest-neighbor search. Performance is benchmarked across two regimes: \emph{intra-sequence} evaluation on the KITTI dataset (sequences 02, 05, and 06) to assess short-term viewpoint robustness (enforcing a 200-frame temporal exclusion window to preclude trivial matches) and \emph{inter-sequence} evaluation on the NCLT dataset to test invariance against long-term seasonal and appearance variations. In the latter, the \textit{2012-01-15} traversal serves as the database against four multi-seasonal query sequences. A revisit is registered if the Euclidean distance $d_q$ between descriptors satisfies a decision threshold $\tau=5$, with results quantified through $\text{Recall}@\{1,5,10,20\}$, the maximum $F_1$ score, and PR AUC, and MMR. To ensure metric integrity, we filter queries lacking pose-based ground-truth positives, focusing strictly on retrieval efficacy in valid revisit scenarios.

\section{Results and Discussion}

This section presents an evaluation of the considered lightweight architectures under different precision regimes.

\subsection{Quantitative Results}

\begin{figure*}[t!]
    \centering
    \includegraphics[width=\linewidth]{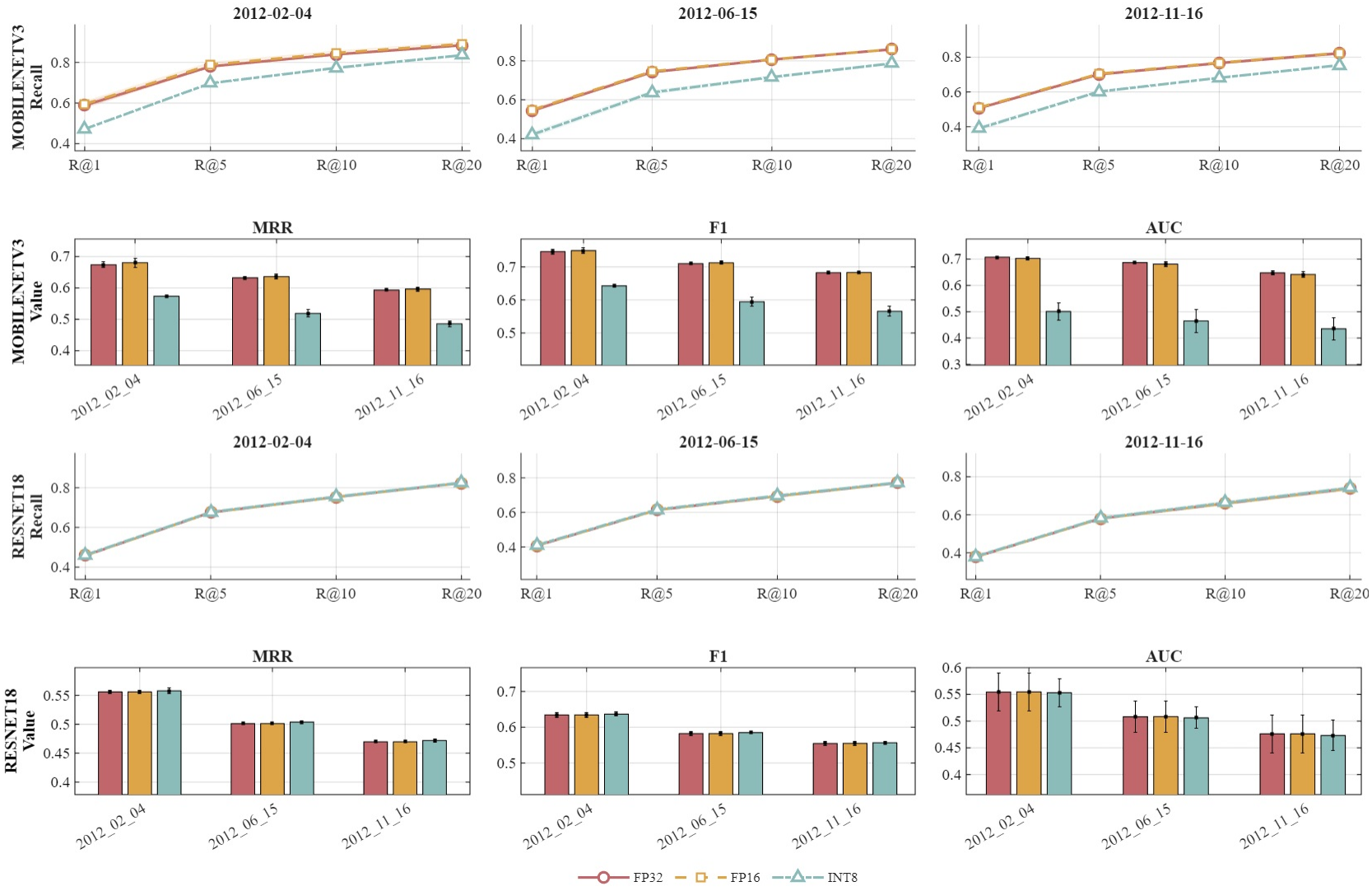}
    \caption{An example of different quantization performance (FP32, FP16, and INT8) on two models   across multiple sequences for MobileNetv2 and ResNet18. Results are aggregated across three experimental seeds, showing the mean and standard deviation for key metrics including Recall@K, MRR, F1-Score, and AUC. Black lines in the bar plot highlight the highest and the lowest values. INT8 quantization reveals varying degrees of performance degradation for MobileNetv2.}
    \label{fig:ncltresults}
\end{figure*}

Table~\ref{tab:model_specs} reports model complexity and memory footprint, while performance is evaluated across multiple precision formats (see \ref{subsection:quantization}) as in Figure~\ref{fig:ncltresults}. Two main trends emerge. First, FP32 and FP16 yield nearly identical results across all architectures, indicating that half-precision provides a “free” gain in efficiency without affecting accuracy. Second, the impact of INT8 quantization is architecture-dependent: MobileNetV3 shows a clear degradation (Recall@1 from $\sim0.74$ to $\sim0.65$), whereas ShuffleNetV2 remains largely stable, and ResNet18 exhibits strong robustness, with negligible performance degradation. Dataset characteristics further influence these trends. On KITTI, where temporal variation is limited, all models perform consistently well. MobileNetV3 achieves the highest accuracy, followed by ShuffleNetV2 and ResNet18, with only minor degradation under INT8. On NCLT, which introduces long-term appearance changes (months to one year), performance drops for all models. In this more challenging setting, ResNet18 shows improved robustness, particularly under quantization, suggesting better generalization to temporal variations. In contrast, MobileNetV3 is more sensitive to INT8 compression, while ShuffleNetV2 maintains stable but lower performance. These results highlight a trade-off between efficiency and robustness in long-term place recognition. 

\subsection{Architecture vs Efficiency Trade-offs}

When analyzing performance jointly with model size, clear trade-offs emerge. ResNet18 achieves the highest robustness to quantization but comes at a significant cost in memory (44.6 MB in FP32), making it less suitable for deployment on constrained platforms. On the opposite end, ShuffleNetV2 offers the most favorable balance, requiring only 8.69 MB in FP32 and 2.17 MB in INT8, while maintaining competitive performance. MobileNetV3 occupies an intermediate position, achieving the best overall accuracy in floating-point precision but suffering the largest degradation under INT8 compression. A further observation emerges from Table~\ref{tab:inference} when considering inference behavior: lightweight architectures do not necessarily translate to higher efficiency under all deployment conditions. In particular, despite its reduced parameter count, ShuffleNetV2 exhibits limited throughput scalability under INT8 CPU execution, suggesting that architectural efficiency in terms of parameters does not directly imply efficient hardware utilization.


\subsection{Implications for Edge Deployment}

From an EdgeAI perspective, these results suggest that the optimal architecture depends on the target constraints. If maximum accuracy is required and moderate resources are available, MobileNetV3 in FP16 represents a strong candidate. However, for strict memory and power budgets, ShuffleNetV2 with INT8 quantization provides the most efficient solution, achieving competitive performance with a $\sim4\times$ reduction in memory compared to FP32. The robustness of ResNet18 to INT8 quantization is an interesting finding, but its larger footprint limits its applicability in low-power scenarios. This highlights an important design insight: robustness to quantization alone is insufficient if not paired with architectural efficiency.

\input{results_table}

\subsection{Discussion}

The results underline that lightweight architectures can effectively support LiDAR place recognition when combined with BEV representations and descriptor-based retrieval. Models optimized for mobile vision (e.g., MobileNet) are not necessarily optimal under aggressive quantization, while architectures with residual connections appear more stable. \emph{These findings emphasize the importance of co-designing architecture, descriptor dimensionality, and numerical precision when targeting real-world autonomous systems.} In particular, the ability to maintain high retrieval performance under INT8 constraints is a key enabler for scalable, long-term deployment on low-power robotic platforms.

\section{Conclusion}

This work investigated the design of efficient LiDAR-based place recognition systems under resource constraints, with a focus on lightweight architectures and low-precision inference. By leveraging BEV representations, we enabled the use of compact image-based models and conducted a systematic comparison of MobileNetV3, ShuffleNetV2, and ResNet18 under FP32, FP16, and INT8 settings. The results highlight that FP16 offers an effective trade-off between efficiency and accuracy, while INT8 quantization introduces architecture-dependent effects. In particular, ShuffleNetV2 provides the best balance between performance and resource consumption, whereas ResNet18 demonstrates higher robustness to quantization and long-term appearance changes. These findings emphasize that both architectural design and numerical precision must be jointly considered when targeting deployment on low-power platforms. Future work will explore joint optimization strategies, including quantization-aware training and descriptor dimensionality reduction, as well as the extension to multi-modal perception pipelines. Ultimately, this study provides practical guidelines for the development of scalable and efficient place recognition systems in real-world autonomous applications. Future work will explore joint optimization strategies, including quantization-aware training and descriptor dimensionality reduction. In addition, a finer-grained layer-wise quantization analysis could be investigated to better understand the sensitivity of specific operations (particularly non-linear activations) to low-precision representations.

\balance
\bibliography{bib}
\bibliographystyle{IEEEtran}
\

\end{document}

%% file: results_table.tex
\begin{table*}[t]
\centering
\caption{Performance comparison on the KITTI dataset (decimals). Values are averaged across the dataset sequences}
\label{tab:kitti_results}
\small
\setlength{\tabcolsep}{3.5pt}
\begin{tabular}{lcccccccccccc}
\toprule
& \multicolumn{4}{c}{\textbf{FP32}} 
& \multicolumn{4}{c}{\textbf{FP16}} 
& \multicolumn{4}{c}{\textbf{INT8}} \\
\cmidrule(lr){2-5} \cmidrule(lr){6-9} \cmidrule(lr){10-13}
\textbf{Model}
& R@1 & R@5 & F1 & AUC
& R@1 & R@5 & F1 & AUC
& R@1 & R@5 & F1 & AUC \\
\midrule

MobileNetV3 
& \textbf{0.902} & \textbf{0.945} & \textbf{0.889} & \textbf{0.917}
& \textbf{0.895} & \textbf{0.940} & \textbf{0.880} & 0.900
& 0.750 & 0.800 & 0.730 & 0.742 \\

ResNet18
& 0.845 & 0.910 & 0.830 & 0.890
& 0.846 & 0.910 & 0.831 & 0.890
& 0.848 & 0.912 & 0.835 & \textbf{0.895} \\

ShuffleNetV2
& 0.880 & 0.935 & 0.870 & 0.905
& 0.880 & 0.935 & 0.870 & \textbf{0.905}
& \textbf{0.870} & \textbf{0.930} & \textbf{0.865} & 0.890 \\

\bottomrule
\end{tabular}
\end{table*}

\begin{table*}[t]
\centering
\caption{Performance comparison on the NCLT dataset (decimals). }
\label{tab:nclt_results}
\small
\setlength{\tabcolsep}{3.5pt}
\begin{tabular}{lcccccccccccc}
\toprule
& \multicolumn{4}{c}{\textbf{FP32}} 
& \multicolumn{4}{c}{\textbf{FP16}} 
& \multicolumn{4}{c}{\textbf{INT8}} \\
\cmidrule(lr){2-5} \cmidrule(lr){6-9} \cmidrule(lr){10-13}
\textbf{Model}
& R@1 & R@5 & F1 & AUC
& R@1 & R@5 & F1 & AUC
& R@1 & R@5 & F1 & AUC \\
\midrule

MobileNetV3 
& \textbf{0.582} & \textbf{0.763} & \textbf{0.610} & \textbf{0.723}
& \textbf{0.589} & \textbf{0.774} & \textbf{0.615} & \textbf{0.740}
& 0.538 & \textbf{0.785} & 0.575 & 0.618 \\

ResNet18
& 0.465 & 0.656 & 0.500 & 0.562
& 0.466 & 0.656 & 0.501 & 0.562
& 0.466 & 0.660 & 0.505 & 0.565 \\

ShuffleNetV2
& 0.548 & 0.738 & 0.580 & 0.679
& 0.548 & 0.738 & 0.580 & 0.679
& \textbf{0.544} & 0.738 & \textbf{0.578} & \textbf{0.672} \\

\bottomrule
\end{tabular}
\end{table*}

\begin{table}[t]
\centering
\caption{Inference performance in terms of latency and throughput, for batch sizes 1 and 32. FP32/FP16 are evaluated on GPU, while INT8 is executed on CPU.}
\label{tab:inference}
\resizebox{\columnwidth}{!}{
\begin{tabular}{llcccc}
\toprule
Model & Prec. & \multicolumn{2}{c}{Batch = 1} & \multicolumn{2}{c}{Batch = 32} \\
\cmidrule(r){3-4} \cmidrule(l){5-6}
 &  & Lat. [ms] & Thr. [img/s] & Lat. [ms] & Thr. [img/s] \\
\midrule
\multirow{3}{*}{ShuffleNetV2}
 & FP32 & 4.95  & 202.2  & 5.49  & 5831.9 \\
 & FP16 & 6.26  & 159.7  & 6.12  & 5227.7 \\
 & INT8 & 103.85 & 9.6   & 3251.71 & 9.8 \\
\midrule
\multirow{3}{*}{MobileNetV3}
 & FP32 & 4.79  & 208.6  & 10.87 & 2942.8 \\
 & FP16 & 5.62  & 177.9  & 5.83  & 5486.9 \\
 & INT8 & 16.72 & 59.8  & 359.48 & 89.0 \\
\midrule
\multirow{3}{*}{ResNet18}
 & FP32 & 2.13  & 468.6  & 12.41 & 2577.7 \\
 & FP16 & 2.38  & 421.0  & 6.94  & 4609.6 \\
 & INT8 & 25.27 & 39.6  & 781.38 & 41.0 \\
\bottomrule
\end{tabular}%
} 
\end{table}